\pdfoutput=1

\documentclass[11pt]{article}
\usepackage{svg}

\usepackage[final]{acl}

\usepackage{times}
\usepackage{latexsym}
\usepackage{graphicx}
\usepackage{graphicx}
\usepackage{textcomp}
\usepackage{xcolor}
\usepackage{algorithm}
\usepackage{float}
\usepackage{amsmath,amssymb}
\usepackage{subcaption}
\usepackage{url}
\usepackage{hyperref}

\usepackage[T1]{fontenc}

\usepackage[utf8]{inputenc}

\usepackage{microtype}

\usepackage{inconsolata}

\title{Mast Kalandar at SemEval-2024 Task 8: On the Trail of Textual Origins: RoBERTa-BiLSTM Approach to Detect AI-Generated Text}

\author{Jainit Sushil Bafna \\
  IIIT Hyderabad \\
  {\footnotesize \texttt{jainit.bafna@research.iiit.ac.in\ } } \\\And
  Hardik Mittal\thanks{Equal contribution.} \\
  IIIT Hyderabad \\
  {\footnotesize \texttt{\ hardik.mittal@research.iiit.ac.in\ } } \\\AND
  Suyash Sethia\footnotemark[1] \\
  IIIT Hyderabad \\
  {\footnotesize \texttt{\ suyash.sethia@research.iiit.ac.in\ } } \\\And
  Manish Shrivastava \\
  IIIT Hyderabad \\
  {\footnotesize \texttt{\ m.shrivastava@iiit.ac.in} } 
  \\\And
  Radhika Mamidi \\
  IIIT Hyderabad \\
  {\footnotesize \texttt{\ radhika.mamidi@iiit.ac.in} } }

\begin{document}
\maketitle 
\begin{abstract}
Large Language Models (LLMs) have showcased impressive abilities in generating fluent responses to diverse user queries. However, concerns regarding the potential misuse of such texts in journalism, educational, and academic contexts have surfaced. SemEval 2024 introduces the task of Multigenerator, Multidomain, and Multilingual Black-Box Machine-Generated Text Detection, aiming to develop automated systems for identifying machine-generated text and detecting potential misuse. In this paper, we i) propose a RoBERTa-BiLSTM based classifier designed to classify text into two categories: AI-generated or human ii) conduct a comparative study of our model with baseline approaches to evaluate its effectiveness. This paper contributes to the advancement of automatic text detection systems in addressing the challenges posed by machine-generated text misuse. Our architecture ranked 46th on the official leaderboard with an accuracy of 80.83 among 125. 
\end{abstract}

\section{Introduction}
The task of classifying text as either AI-generated or human-generated holds significant importance in the field of natural language processing (NLP). It addresses the growing need to distinguish between content created by artificial intelligence models and that generated by human authors, a distinction crucial for various applications such as content moderation, misinformation detection, and safeguarding against AI-generated malicious content. This task is outlined in the task overview paper by \cite{wang2023m4}, emphasizing its relevance and scope in the NLP community.

Our system employs a hybrid approach combining deep learning techniques with feature engineering to tackle the classification task effectively. Specifically, we leverage a BiLSTM (Bidirectional Long Short-Term Memory) \cite{bilstmarticle} neural network in conjunction with RoBERTa \cite{liu2019roberta}, a pre-trained language representation model, to capture both sequential and contextual information from the input sentences. This hybrid architecture enables our system to effectively capture nuanced linguistic patterns and semantic cues for accurate classification.

Participating in this task provided valuable insights into the capabilities and limitations of our system. Quantitatively, our system achieved competitive results, ranking 46 relative to other teams in terms of accuracy and F1 score. Qualitatively, we observed that our system struggled with distinguishing between sentences generated by AI models trained on specific domains or datasets with highly similar linguistic patterns.

We have released the code for our system on GitHub\footnote{\url{ https://github.com/Mast-Kalandar/SemEval2024-task8}}, facilitating transparency and reproducibility in our approach.

\section{Related Works}

\begin{table*}
\centering
a)
\begin{tabular}{llllll}
\hline
\textbf{Model/Source} & \textbf{chatGPT} & \textbf{cohere} & \textbf{davinci} & \textbf{dolly} & \textbf{human} \\
\hline
wikihow & 3000 & 3000 & 3000 & 3000 & 15499 \\
wikipedia & 2995 & 2336 & 3000 & 2702 & 14497 \\
reddit & 3000 & 3000 & 3000 & 3000 & 15500 \\
arxiv & 3000 & 3000 & 2999 & 3000 & 15498 \\
peerread & 2344 & 2342 & 2344 & 2344 & 2357 \\
\hline
\end{tabular} \\
\vspace{5px}
b)
\begin{tabular}{lcc}
\hline
\textbf{Model/Source} & \textbf{bloomz} & \textbf{human} \\
\hline
wikihow & 500 & 500 \\
wikipedia & 500 & 500 \\
reddit & 500 & 500 \\
arxiv & 500 & 500 \\
peerread & 500 & 500 \\
\hline
\end{tabular}
\caption{Table a) contains statistics about the train split. Table b) contains statistics about the validation split from the M4 dataset}
\label{tab:accents}
\end{table*}
In the field of detecting machine-generated text, numerous methodologies and models have been examined. A distinguished methodology is the application of the RoBERTa Classifier, which enhances the RoBERTa language model through fine-tuning for the specific purpose of identifying machine-generated text. The proficiency of pre-trained classifiers like RoBERTa in this domain has been affirmed through various studies, including those conducted by \cite{DBLP:journals/corr/abs-1908-09203} and additional research by \cite{NEURIPS2019_3e9f0fc9,DBLP:journals/corr/abs-1911-00650,DBLP:journals/corr/abs-2004-10188,app10175841, DBLP:journals/corr/abs-2109-13296}. Concurrently, the XLM-R Classifier exploits the multilingual training of the XLM-RoBERTa model to effectively recognize machine-generated text in various languages, as demonstrated by \cite{DBLP:journals/corr/abs-1911-02116}.

Alternatively, the exploration of logistic regression models that incorporate GLTR (Giant Language model Test Room) features has been undertaken. These models strive to discern subtleties in text generation methodologies by analyzing token probabilities and distribution entropy, as investigated by \cite{gehrmann-etal-2019}. Furthermore, detection efforts have utilized stylometric and NELA (News Landscape) features, which account for a broad spectrum of linguistic and structural characteristics, including syntactic, stylistic, affective, and moral dimensions, as reported by \cite{6982099} and \cite{mitchell2023detectgpt}. Additionally, proprietary frameworks like GPTZero, devised by Princeton University, focus on indicators such as perplexity and burstiness to analyze texts for machine-generated content identification. Although the specific technical details are sparingly disclosed, the reported effectiveness of GPTZero in identifying outputs from various AI language models highlights its significance in the ongoing development of machine-generated text detection strategies \cite{Ouyang2022TrainingLM,NEURIPS2020_1457c0d6, Radford2019LanguageMA, touvron2023_llama}.

\section{Background}
\begin{table*}
\centering
  \begin{tabular}{llllll}
\hline
\textbf{Model} & \textbf{Accuracy} & \textbf{F1} & \textbf{Precision} & \textbf{Recall} & \textbf{Params*} \\
\hline

Full RoBERTa fine tune & {80.68 } & {80.54 } & {81.55 } & {80.68 } & 124M \\
LoRA with RoBERTa (Freezed) & {81.59 } & {81.06 } & {85.64 } & {\textbf{81.59} } & 0.7M \\
LoRA with LongFormer & {75.34 } & {75.14 } & {76.16 } & {75.34 } & 6M \\
BiLSTM with RoBERTa (Un-Freezed) & {70.77 } & {61.15 } & {\textbf{91.19} } & {46.00 } & 18M \\
GRU with RoBERTa (Freezed) & {74.65 } & {80.54 } & {81.55 } & {80.68 } & 3M \\
\textbf{BiLSTM with RoBERTa (Freezed) } & \textbf{82.52 } & \textbf{82.14 } & {83.96 } & {80.40 } & 4M \\
\hline
\end{tabular}
\caption{The performance of the models tried on the dev set of the dataset.\\ *The params only accounts for trainable unfreezed parameters.}
\label{tab:scores}
\end{table*}

\subsection{Dataset}

For the machine-generated text, the researchers used various multilingual language models like ChatGPT\cite{chatgpt}, textdavinci-003\cite{text-davinci-003}, LLaMa\cite{touvron2023llama}, FlanT5\cite{chung2022scaling}, Cohere\cite{cohere}, Dolly-v2\cite{dolly}, and BLOOMz\cite{muennighoff2023crosslingual}. These models were given different tasks like writing Wikipedia articles, summarizing abstracts from arXiv, providing peer reviews, answering questions from Reddit and Baike/Web QA, and creating news briefs.
As evident from Table \ref{tab:accents}, the training set lacks any sentences generated by the Bloomz model, which stands as the sole model represented in the validation set. This deliberate choice ensures a robust assessment of our model's generalization capabilities across all machine-generated outputs, regardless of the specific model generating them. By exposing our model to diverse machine-generated sentences during training, including those from unseen models like Bloomz in the validation set, we aim to evaluate its ability to effectively generalize to novel inputs and make reliable predictions across the spectrum of machine-generated text.

\subsection{Task}
We focused on Subtask-A of the SemEval Task 8 which involves developing a classifier to differentiate between monolingual sentences generated by artificial intelligence (AI) systems and those generated by humans. This classification task is essential for distinguishing the origin of text and understanding whether it was produced by AI models or by human authors.
\subsubsection{Objective}
The primary objective is to build a robust classifier capable of accurately distinguishing between AI-generated and human-generated sentences. The classifier should generalize well across various AI models and domains, ensuring consistent performance regardless of the specific model or domain from which the text originates. 

The goal was to design a model that not only performs this task with high accuracy but also adapts to various AI models and domains. It's crucial for the classifier to accurately identify the origin of sentences, regardless of the technology used to generate them or their subject matter, ensuring broad applicability and effectiveness

    
    
    

\section{System Overview}

\begin{figure*}[ht]
\centering
\includegraphics[width=1.1\textwidth]{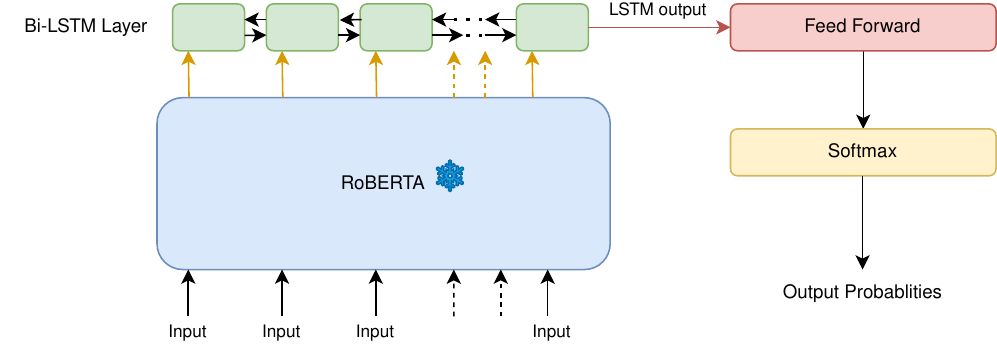}
\caption{Our proposed architecture of BiLSTM with freezed RoBERTa}
\label{fig:svgimage}
\end{figure*}

Based on our observation (See \ref{sec:appendix}), we discovered that language modeling task encodes the various features required for detection of AI written text. So we used pretrained RoBERTa in most of our architectures so exploit this power of language models.
\subsection{Full RoBERTa Finetune}

The Full RoBERTa\cite{liu2019roberta} Finetune model, chosen as our baseline, boasted an extensive architecture and possessed the highest parameter count among the models under evaluation. Serving as a comprehensive starting point, this model allowed us to assess the effectiveness of subsequent enhancements in comparison.

\subsection{LoRA with RoBERTa (Frozen)}

Incorporating Low Rank Adapters \cite{hu2021lora}, we applied fine-tuning techniques to the RoBERTa model while strategically freezing all layers. This approach enabled us to adapt the model to our specific task domain, leveraging pre-trained representations effectively.

\subsection{LoRA with LongFormer}

The limitation of RoBERTa's context length (max 512 tokens) posed challenges for handling lengthy sentences in our dataset. To address this, we investigated LongFormer \cite{beltagy2020longformer}, a model designed to efficiently manage longer contexts. Despite employing LoRA for fine-tuning, the model's performance on the validation set fell short of expectations, indicating potential difficulties in generalization.

\subsection{RoBERTa (2 Layers unfreezed) + BiLSTM}

Expanding upon RoBERTa's capabilities, we introduced a hybrid architecture by unfreezing two layers and integrating a BiLSTM network \cite{bilstmarticle}. RoBERTa served as the primary encoder for sentence representations, with the subsequent BiLSTM layer trained to classify based on the last hidden state. 

\subsection{RoBERTa (Frozen) + GRU}

In our endeavor to augment RoBERTa's capabilities, we devised a hybrid architecture by integrating a Gated Recurrent Unit (GRU) \cite{chung2014empirical} network with the frozen RoBERTa model. Within this framework, RoBERTa served as the encoder for generating sentence representations, while a subsequent GRU layer was incorporated for sequential processing and classification tasks. This amalgamation aimed to leverage the strengths of both RoBERTa's contextual understanding and GRU's recurrent dynamics, contributing to enhanced performance on our target task.

\subsection{RoBERTa (Frozen) + BiLSTM }

In our pursuit of enhancing RoBERTa's capabilities, we devised a hybrid architecture by coupling a Bidirectional Long Short-Term Memory (BiLSTM) network with the RoBERTa model \cite{liu2019roberta}. In this setup, RoBERTa functioned as the encoder for sentence representations, while a subsequent BiLSTM layer was employed for classification, utilizing the last hidden state for decision-making. For a detailed visual representation of the model's architecture, please refer to the accompanying Figure \ref{fig:svgimage}.

\vspace{4px}

We explored various methodologies (refer to Table~\ref{tab:scores} for detailed performance metrics) before selecting the optimal approach as our final model. Subsequently, we assessed the performance of the chosen model, RoBERTA (Freezed) + BiLSTM, on the test dataset.

\section{Experiments}

\subsection{Preprocessing}

All textual data underwent standard preprocessing steps, including tokenization, lowercasing, and punctuation marks. Additionally, specific domain-related preprocessing, such as handling special characters or domain-specific terms, was performed as necessary.

\subsection{Hyperparameter Tuning}

Hyperparameters were tuned using a combination of grid search and random search techniques. We explored various hyperparameter combinations to identify the optimal configuration for each model variant. \\
The configuration for LSTM and GRU used in Table \ref{tab:scores} is \texttt{hidden\_size=256}, \texttt{layers=2}, \texttt{dropout=0.2}, with \texttt{LoRA rank being 20} has been found as the best configuration for the models.
For RoBERTa+LSTM model's feedforward had a single weight matrix of dimension 512*2.








\section{Results}

\begin{table*}[htbp]
\centering
    \begin{tabular}{llllll}
\hline
\textbf{Model} & \textbf{Accuracy} & \textbf{F1} & \textbf{Precision} & \textbf{Recall} & \textbf{Params*} \\
\hline

Full RoBERTa fine tune$^+$ & \textbf{88.47} & \textbf{88.44} & \textbf{93.36} & 84.02 & 124M \\

LoRA with RoBERTa (Freezed) & 80.91 & 80.18 & 83.88 & 80.14 & 0.7M \\

LoRA with LongFormer & 63.39 & 57.51 & 72.45 & 61.67 & 6M \\

BiLSTM with RoBERTa (Un-Freezed) & 80.80 & 80.19 & 83.08 & 80.12 & 18M \\

GRU with RoBERTa (Freezed) & 84.71 & 84.33 & 86.53 & 84.13 & 3M \\

BiLSTM with RoBERTa (Freezed)  &  80.83  & 80.83 &  74.65 & \textbf{96.16} & 4M \\

\hline 
\end{tabular}
\caption{The performance of the models tried on the test set of the dataset.\\ * The params only accounts for trainable unfreezed parameters.\\+ Baseline mentioned in task overview paper}

\label{tab:results}
\end{table*}
We tested our models on various models on the test set. The results can be viewed in (Table: \ref{tab:results}).\\
\textbf{Ranking:} Our BiLSTM+RoBERTa model achieved a ranking of 46 out of 125 participants in the competition, demonstrating its competitive performance (as shown in Table~\ref{tab:results}). These results highlight the effectiveness of various models, including BiLSTM+RoBERTa and GRU+RoBERTa, in addressing the task objectives. We submitted BiLSTM+RoBERTa based on its strong performance on the validation set. However, after testing all models listed in Table~\ref{tab:results}, we found that GRU+RoBERTa achieved a significantly better result, with an accuracy increase of approximately 4\%.

\section{Conclusion}
In conclusion, our BiLSTM+RoBERTa model effectively tackled the task, achieving competitive results, thanks to its deep learning and pre-trained language model. While a similar model with unfrozen RoBERTa boasted higher precision, its complexity came at the cost of increased parameters.

Impressively, our model ranked 46th out of 125 competition entries (Table~\ref{tab:results}), showcasing its potential alongside approaches like GRU+RoBERTa. Interestingly, post-competition analysis revealed GRU+RoBERTa's superior accuracy (by about 4\%). This highlights the value of exploring diverse architectures and hyperparameter tuning for peak performance.

Moving forward, there are several avenues for future work to explore. Firstly, further experimentation with different model architectures, including alternative combinations of encoders and classifiers, could potentially yield improvements in performance. Additionally, fine-tuning hyperparameters and exploring advanced techniques for model optimization may enhance the robustness and generalization capabilities of our system. Furthermore, incorporating additional contextual information or domain-specific knowledge could potentially augment the model's understanding and performance on specific tasks. Overall, our findings contribute to the ongoing research efforts in natural language processing and provide valuable insights for future developments in this domain.

\bibliography{anthology,custom}

\begin{thebibliography}{27}
\expandafter\ifx\csname natexlab\endcsname\relax\def\natexlab#1{#1}\fi

\bibitem[{Bakhtin et~al.(2020)Bakhtin, Deng, Gross, Ott, Ranzato, and Szlam}]{DBLP:journals/corr/abs-2004-10188}
Anton Bakhtin, Yuntian Deng, Sam Gross, Myle Ott, Marc'Aurelio Ranzato, and Arthur Szlam. 2020.
\newblock \href {http://arxiv.org/abs/2004.10188} {Energy-based models for text}.
\newblock \emph{CoRR}, abs/2004.10188.

\bibitem[{Beltagy et~al.(2020)Beltagy, Peters, and Cohan}]{beltagy2020longformer}
Iz~Beltagy, Matthew~E. Peters, and Arman Cohan. 2020.
\newblock \href {http://arxiv.org/abs/2004.05150} {Longformer: The long-document transformer}.

\bibitem[{Brown et~al.(2020)Brown, Mann, Ryder, Subbiah, Kaplan, Dhariwal, Neelakantan, Shyam, Sastry, Askell, Agarwal, Herbert-Voss, Krueger, Henighan, Child, Ramesh, Ziegler, Wu, Winter, Hesse, Chen, Sigler, Litwin, Gray, Chess, Clark, Berner, McCandlish, Radford, Sutskever, and Amodei}]{NEURIPS2020_1457c0d6}
Tom Brown, Benjamin Mann, Nick Ryder, Melanie Subbiah, Jared~D Kaplan, Prafulla Dhariwal, Arvind Neelakantan, Pranav Shyam, Girish Sastry, Amanda Askell, Sandhini Agarwal, Ariel Herbert-Voss, Gretchen Krueger, Tom Henighan, Rewon Child, Aditya Ramesh, Daniel Ziegler, Jeffrey Wu, Clemens Winter, Chris Hesse, Mark Chen, Eric Sigler, Mateusz Litwin, Scott Gray, Benjamin Chess, Jack Clark, Christopher Berner, Sam McCandlish, Alec Radford, Ilya Sutskever, and Dario Amodei. 2020.
\newblock \href {https://proceedings.neurips.cc/paper_files/paper/2020/file/1457c0d6bfcb4967418bfb8ac142f64a-Paper.pdf} {Language models are few-shot learners}.
\newblock In \emph{Advances in Neural Information Processing Systems}, volume~33, pages 1877--1901. Curran Associates, Inc.

\bibitem[{Chung et~al.(2022)Chung, Hou, Longpre, Zoph, Tay, Fedus, Li, Wang, Dehghani, Brahma, Webson, Gu, Dai, Suzgun, Chen, Chowdhery, Castro-Ros, Pellat, Robinson, Valter, Narang, Mishra, Yu, Zhao, Huang, Dai, Yu, Petrov, Chi, Dean, Devlin, Roberts, Zhou, Le, and Wei}]{chung2022scaling}
Hyung~Won Chung, Le~Hou, Shayne Longpre, Barret Zoph, Yi~Tay, William Fedus, Yunxuan Li, Xuezhi Wang, Mostafa Dehghani, Siddhartha Brahma, Albert Webson, Shixiang~Shane Gu, Zhuyun Dai, Mirac Suzgun, Xinyun Chen, Aakanksha Chowdhery, Alex Castro-Ros, Marie Pellat, Kevin Robinson, Dasha Valter, Sharan Narang, Gaurav Mishra, Adams Yu, Vincent Zhao, Yanping Huang, Andrew Dai, Hongkun Yu, Slav Petrov, Ed~H. Chi, Jeff Dean, Jacob Devlin, Adam Roberts, Denny Zhou, Quoc~V. Le, and Jason Wei. 2022.
\newblock \href {http://arxiv.org/abs/2210.11416} {Scaling instruction-finetuned language models}.

\bibitem[{Chung et~al.(2014)Chung, Gulcehre, Cho, and Bengio}]{chung2014empirical}
Junyoung Chung, Caglar Gulcehre, KyungHyun Cho, and Yoshua Bengio. 2014.
\newblock \href {http://arxiv.org/abs/1412.3555} {Empirical evaluation of gated recurrent neural networks on sequence modeling}.

\bibitem[{{Cohere}(2024)}]{cohere}
{Cohere}. 2024.
\newblock {Cohere: Chat}.
\newblock \url{https://cohere.com/}.
\newblock Accessed: February 20, 2024.

\bibitem[{Conneau et~al.(2019)Conneau, Khandelwal, Goyal, Chaudhary, Wenzek, Guzm{\'{a}}n, Grave, Ott, Zettlemoyer, and Stoyanov}]{DBLP:journals/corr/abs-1911-02116}
Alexis Conneau, Kartikay Khandelwal, Naman Goyal, Vishrav Chaudhary, Guillaume Wenzek, Francisco Guzm{\'{a}}n, Edouard Grave, Myle Ott, Luke Zettlemoyer, and Veselin Stoyanov. 2019.
\newblock \href {http://arxiv.org/abs/1911.02116} {Unsupervised cross-lingual representation learning at scale}.
\newblock \emph{CoRR}, abs/1911.02116.

\bibitem[{{databricks}(2022)}]{dolly}
{databricks}. 2022.
\newblock {Free Dolly: Introducing the World's First Truly Open Instruction-Tuned LLM}.
\newblock \href{https://www.databricks.com/blog/2023/04/12 /dolly-first-open-commercially-viable-instruction-tuned-llm}{dolly-v2}.
\newblock Accessed: February 20, 2024.

\bibitem[{Gehrmann et~al.(2019)Gehrmann, Strobelt, and Rush}]{gehrmann-etal-2019}
Sebastian Gehrmann, Hendrik Strobelt, and Alexander Rush. 2019.
\newblock \href {https://doi.org/10.18653/v1/P19-3019} {{GLTR}: Statistical detection and visualization of generated text}.
\newblock In \emph{Proceedings of the 57th Annual Meeting of the Association for Computational Linguistics: System Demonstrations}, pages 111--116, Florence, Italy. Association for Computational Linguistics.

\bibitem[{Hu et~al.(2021)Hu, Shen, Wallis, Allen-Zhu, Li, Wang, Wang, and Chen}]{hu2021lora}
Edward~J. Hu, Yelong Shen, Phillip Wallis, Zeyuan Allen-Zhu, Yuanzhi Li, Shean Wang, Lu~Wang, and Weizhu Chen. 2021.
\newblock \href {http://arxiv.org/abs/2106.09685} {Lora: Low-rank adaptation of large language models}.

\bibitem[{Ippolito et~al.(2019)Ippolito, Duckworth, Callison{-}Burch, and Eck}]{DBLP:journals/corr/abs-1911-00650}
Daphne Ippolito, Daniel Duckworth, Chris Callison{-}Burch, and Douglas Eck. 2019.
\newblock \href {http://arxiv.org/abs/1911.00650} {Human and automatic detection of generated text}.
\newblock \emph{CoRR}, abs/1911.00650.

\bibitem[{Jang et~al.(2020)Jang, Kim, Harerimana, Kang, and Kim}]{app10175841}
Beakcheol Jang, Myeonghwi Kim, Gaspard Harerimana, Sang-ug Kang, and Jong~Wook Kim. 2020.
\newblock \href {https://doi.org/10.3390/app10175841} {Bi-lstm model to increase accuracy in text classification: Combining word2vec cnn and attention mechanism}.
\newblock \emph{Applied Sciences}, 10(17).

\bibitem[{Li et~al.(2014)Li, Monaco, Chen, and Tappert}]{6982099}
Jenny~S. Li, John~V. Monaco, Li-Chiou Chen, and Charles~C. Tappert. 2014.
\newblock \href {https://doi.org/10.1109/ICEBE.2014.61} {Authorship authentication using short messages from social networking sites}.
\newblock In \emph{2014 IEEE 11th International Conference on e-Business Engineering}, pages 314--319.

\bibitem[{Liu et~al.(2019)Liu, Ott, Goyal, Du, Joshi, Chen, Levy, Lewis, Zettlemoyer, and Stoyanov}]{liu2019roberta}
Yinhan Liu, Myle Ott, Naman Goyal, Jingfei Du, Mandar Joshi, Danqi Chen, Omer Levy, Mike Lewis, Luke Zettlemoyer, and Veselin Stoyanov. 2019.
\newblock \href {http://arxiv.org/abs/1907.11692} {Roberta: A robustly optimized bert pretraining approach}.

\bibitem[{Mitchell et~al.(2023)Mitchell, Lee, Khazatsky, Manning, and Finn}]{mitchell2023detectgpt}
Eric Mitchell, Yoonho Lee, Alexander Khazatsky, Christopher~D. Manning, and Chelsea Finn. 2023.
\newblock \href {http://arxiv.org/abs/2301.11305} {Detectgpt: Zero-shot machine-generated text detection using probability curvature}.

\bibitem[{Muennighoff et~al.(2023)Muennighoff, Wang, Sutawika, Roberts, Biderman, Scao, Bari, Shen, Yong, Schoelkopf, Tang, Radev, Aji, Almubarak, Albanie, Alyafeai, Webson, Raff, and Raffel}]{muennighoff2023crosslingual}
Niklas Muennighoff, Thomas Wang, Lintang Sutawika, Adam Roberts, Stella Biderman, Teven~Le Scao, M~Saiful Bari, Sheng Shen, Zheng-Xin Yong, Hailey Schoelkopf, Xiangru Tang, Dragomir Radev, Alham~Fikri Aji, Khalid Almubarak, Samuel Albanie, Zaid Alyafeai, Albert Webson, Edward Raff, and Colin Raffel. 2023.
\newblock \href {http://arxiv.org/abs/2211.01786} {Crosslingual generalization through multitask finetuning}.

\bibitem[{{OpenAI}(2022)}]{text-davinci-003}
{OpenAI}. 2022.
\newblock {text-davinci-003: A Variant of the GPT-3 Language Model}.
\newblock \url{https://openai.com}.
\newblock Accessed: February 20, 2024.

\bibitem[{{OpenAI}(2024)}]{chatgpt}
{OpenAI}. 2024.
\newblock {ChatGPT: A Large-Scale Transformer-Based Language Model}.
\newblock \url{https://openai.com/research/chatgpt}.
\newblock Accessed: February 20, 2024.

\bibitem[{Ouyang et~al.(2022)Ouyang, Wu, Jiang, Almeida, Wainwright, Mishkin, Zhang, Agarwal, Slama, Ray, Schulman, Hilton, Kelton, Miller, Simens, Askell, Welinder, Christiano, Leike, and Lowe}]{Ouyang2022TrainingLM}
Long Ouyang, Jeff Wu, Xu~Jiang, Diogo Almeida, Carroll~L. Wainwright, Pamela Mishkin, Chong Zhang, Sandhini Agarwal, Katarina Slama, Alex Ray, John Schulman, Jacob Hilton, Fraser Kelton, Luke~E. Miller, Maddie Simens, Amanda Askell, Peter Welinder, Paul~Francis Christiano, Jan Leike, and Ryan~J. Lowe. 2022.
\newblock \href {https://api.semanticscholar.org/CorpusID:246426909} {Training language models to follow instructions with human feedback}.
\newblock \emph{ArXiv}, abs/2203.02155.

\bibitem[{Radford et~al.(2019)Radford, Wu, Child, Luan, Amodei, and Sutskever}]{Radford2019LanguageMA}
Alec Radford, Jeff Wu, Rewon Child, David Luan, Dario Amodei, and Ilya Sutskever. 2019.
\newblock \href {https://api.semanticscholar.org/CorpusID:160025533} {Language models are unsupervised multitask learners}.
\newblock In \emph{Language Models are Unsupervised Multitask Learners}.

\bibitem[{Schuster and Paliwal(1997)}]{bilstmarticle}
Mike Schuster and Kuldip Paliwal. 1997.
\newblock \href {https://doi.org/10.1109/78.650093} {Bidirectional recurrent neural networks}.
\newblock \emph{Signal Processing, IEEE Transactions on}, 45:2673 -- 2681.

\bibitem[{Solaiman et~al.(2019)Solaiman, Brundage, Clark, Askell, Herbert{-}Voss, Wu, Radford, and Wang}]{DBLP:journals/corr/abs-1908-09203}
Irene Solaiman, Miles Brundage, Jack Clark, Amanda Askell, Ariel Herbert{-}Voss, Jeff Wu, Alec Radford, and Jasmine Wang. 2019.
\newblock \href {http://arxiv.org/abs/1908.09203} {Release strategies and the social impacts of language models}.
\newblock \emph{CoRR}, abs/1908.09203.

\bibitem[{Touvron et~al.(2023{\natexlab{a}})Touvron, Lavril, Izacard, Martinet, Lachaux, Lacroix, Rozière, Goyal, Hambro, Azhar, Rodriguez, Joulin, Grave, and Lample}]{touvron2023_llama}
Hugo Touvron, Thibaut Lavril, Gautier Izacard, Xavier Martinet, Marie-Anne Lachaux, Timothée Lacroix, Baptiste Rozière, Naman Goyal, Eric Hambro, Faisal Azhar, Aurelien Rodriguez, Armand Joulin, Edouard Grave, and Guillaume Lample. 2023{\natexlab{a}}.
\newblock \href {http://arxiv.org/abs/2302.13971} {Llama: Open and efficient foundation language models}.

\bibitem[{Touvron et~al.(2023{\natexlab{b}})Touvron, Lavril, Izacard, Martinet, Lachaux, Lacroix, Rozière, Goyal, Hambro, Azhar, Rodriguez, Joulin, Grave, and Lample}]{touvron2023llama}
Hugo Touvron, Thibaut Lavril, Gautier Izacard, Xavier Martinet, Marie-Anne Lachaux, Timothée Lacroix, Baptiste Rozière, Naman Goyal, Eric Hambro, Faisal Azhar, Aurelien Rodriguez, Armand Joulin, Edouard Grave, and Guillaume Lample. 2023{\natexlab{b}}.
\newblock \href {http://arxiv.org/abs/2302.13971} {Llama: Open and efficient foundation language models}.

\bibitem[{Uchendu et~al.(2021)Uchendu, Ma, Le, Zhang, and Lee}]{DBLP:journals/corr/abs-2109-13296}
Adaku Uchendu, Zeyu Ma, Thai Le, Rui Zhang, and Dongwon Lee. 2021.
\newblock \href {http://arxiv.org/abs/2109.13296} {{TURINGBENCH:} {A} benchmark environment for turing test in the age of neural text generation}.
\newblock \emph{CoRR}, abs/2109.13296.

\bibitem[{Wang et~al.(2024)Wang, Mansurov, Ivanov, Su, Shelmanov, Tsvigun, Whitehouse, Afzal, Mahmoud, Sasaki, Arnold, Aji, Habash, Gurevych, and Nakov}]{wang2023m4}
Yuxia Wang, Jonibek Mansurov, Petar Ivanov, Jinyan Su, Artem Shelmanov, Akim Tsvigun, Chenxi Whitehouse, Osama~Mohammed Afzal, Tarek Mahmoud, Toru Sasaki, Thomas Arnold, Alham~Fikri Aji, Nizar Habash, Iryna Gurevych, and Preslav Nakov. 2024.
\newblock M4: Multi-generator, multi-domain, and multi-lingual black-box machine-generated text detection.
\newblock In \emph{Proceedings of the 18th Conference of the European Chapter of the Association for Computational Linguistics}, Malta.

\bibitem[{Zellers et~al.(2019)Zellers, Holtzman, Rashkin, Bisk, Farhadi, Roesner, and Choi}]{NEURIPS2019_3e9f0fc9}
Rowan Zellers, Ari Holtzman, Hannah Rashkin, Yonatan Bisk, Ali Farhadi, Franziska Roesner, and Yejin Choi. 2019.
\newblock \href {https://proceedings.neurips.cc/paper_files/paper/2019/file/3e9f0fc9b2f89e043bc6233994dfcf76-Paper.pdf} {Defending against neural fake news}.
\newblock In \emph{Advances in Neural Information Processing Systems}, volume~32. Curran Associates, Inc.

\end{thebibliography}
\clearpage



\newpage
\section*{Appendix A}
\label{sec:appendix}

\subsection*{A. Setup}
In this study, we implemented a methodology aimed at distinguishing human-generated sentences from machine-generated ones within a training dataset. To achieve this, we initially segregated the dataset into two distinct subsets: one containing human-generated sentences and the other comprising machine-generated ones. Subsequently, we trained separate models utilizing these segregated datasets. Specifically, we employed two distinct models for this task : i) Bidirectional Long Short-Term Memory (\textbf{BiLSTM}) model,
ii) \textbf{RoBERTa} model.

Following the training phase, we proceeded to evaluate the performance of both models on a validation dataset. During this evaluation, we measured the loss incurred by each model when tasked with discerning between human-generated and machine-generated sentences. This evaluation process was crucial for assessing the efficacy and generalization capabilities of the trained models in accurately distinguishing between the two types of sentences.
\subsection*{B. Results}
\begin{figure}[H]
    \centering
    \begin{subfigure}[b]{0.45\linewidth}
        \includegraphics[width=\linewidth]{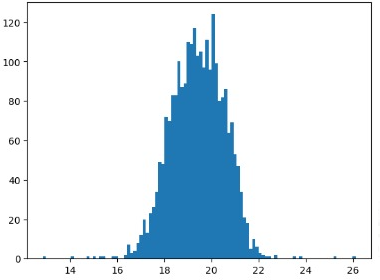}
        \caption{Model Trained on : Human Sentences, Losses Computed on : Human Sentences}
        \label{fig:sub1}
    \end{subfigure}
    \hfill
    \begin{subfigure}[b]{0.45\linewidth}
        \includegraphics[width=\linewidth]{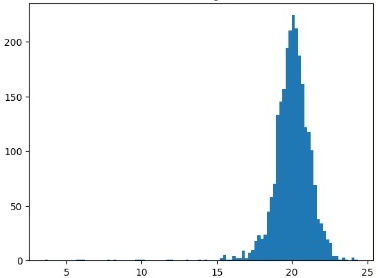}
        \caption{Model Trained on : Human Sentences, Losses Computed on : Machine Sentences}
        \label{fig:sub2}
    \end{subfigure}

    \medskip

    \begin{subfigure}[b]{0.45\linewidth}
        \includegraphics[width=\linewidth]{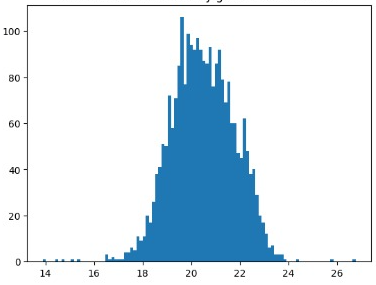}
        \caption{Model Trained on : Machine Sentences, Losses Computed on : Human Sentences}
        \label{fig:sub3}
    \end{subfigure}
    \hfill
    \begin{subfigure}[b]{0.45\linewidth}
        \includegraphics[width=\linewidth]{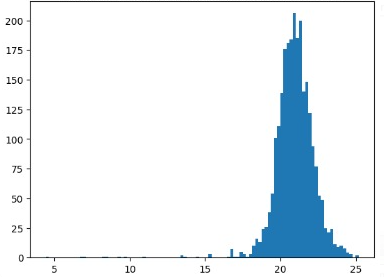}
        \caption{Model Trained on : Machine Sentences, Losses Computed on : Machine Sentences}
        \label{fig:sub4}
    \end{subfigure}

    \caption{Overall Results on Models trained on Human and Machine Generated Sentences and Losses Calculated on Human and Machine Generated Sentences}
    \label{fig:all_images}
\end{figure}

The results are in form of graphs in Figure \ref{fig:all_images}

We noted a consistent pattern across both sets of models – those trained on human-generated sentences and those trained on machine-generated sentences. Specifically, we observed that the losses incurred by human-generated sentences on the validation set exhibited a wider distribution with higher variance, while the losses associated with machine-generated sentences displayed a narrower distribution with lesser variance.

This observation leads to a compelling inference regarding the predictive nature of the model losses for each type of data. The wider distribution and higher variance in losses for human-generated sentences suggest a greater level of unpredictability associated with these sentences. In contrast, the narrower distribution and lesser variance in losses for machine-generated sentences indicate a higher level of predictiveness in the model's performance on these sentences.

This finding sheds light on the inherent characteristics of human-generated versus machine-generated sentences, particularly regarding their predictability when processed by the trained models. Such insights are crucial for understanding the intricacies of model behavior and the challenges posed by different types of data in natural language processing tasks.

\end{document}